\documentclass[runningheads]{llncs}

\usepackage[T1]{fontenc}
\usepackage{graphicx}
\usepackage{silence}
\usepackage{amsmath, amssymb}
\usepackage{booktabs}
\usepackage{multirow}
\usepackage{hyperref}
\usepackage{xcolor}
\newcommand{\rev}[1]{#1}
\begin{document}
	
	% -------------------------------
	% Title
	% -------------------------------
	\title{From Alignment to Prediction: A Study of Self-Supervised Learning and Predictive Representation Learning}
	\titlerunning{Self-Supervised Learning }
	
	% -------------------------------
	% Authors
	% -------------------------------
	\author{Mintu Dutta\inst{1}, Ritesh Vyas\inst{1}, \and Mohendra Roy\inst{1, 2}}
	
	\authorrunning{M. Dutta et al.}
	
	\institute{
		Department of Information and Communication Technology, School of Technology,
		Pandit Deendayal Energy University, Gandhinagar, Gujarat 382007, India \\
		%\email{mintu.dutta@ieee.org}
		\and
		Corresponding Author: mohendra.roy@ieee.org
	}
	
	\maketitle
	% -------------------------------
	% Abstract
	% -------------------------------
	\begin{abstract}
Self-supervised learning has emerged as a major technique for
the task of learning from unlabeled data, where the current methods
mostly revolve around alignment of representations and input recon struction. Although such approaches have demonstrated excellent performance in practice, their scope remains mostly confined to learning
from observed data and do not provide much help in terms of learning structure that is predictive of the data distribution. In this paper, we study some of the recent developments in the realm of self-
supervised learning. We define a new category called Predictive Representation Learning (PRL) which revolves around latent prediction
of unobserved components of data based on the observation. We propose a common taxonomy that classifies PRL along with alignment
and reconstruction-based learning approaches. Furthermore, we argue
that Joint-Embedding Predictive Architecture(JEPA) can be considered as an exemplary member of this new
paradigm. We further discuss theoretical perspectives and open challenges, highlighting predictive representation learning as a promising direction for future self-supervised learning research. In this study, we implemented Bootstrap Your Own Latent (BYOL), Masked Autoencoders (MAE), and Image-JEPA (I-JEPA) for comparative analysis. The results indicate that MAE achieves perfect similarity of 1.00, but exhibits relatively weak robustness of 0.55. In contrast, BYOL and I-JEPA attain accuracies of 0.98 and 0.95, with robustness scores of 0.75 and 0.78, respectively.

		\keywords{Self-Supervised Learning, Predictive Representation Learning, JEPA, Representation Learning}
	\end{abstract}
	
	% =====================================================
	\section{Introduction}
	% =====================================================
	Self-supervised learning has revolutionized representation learning by enabling models to learn from vast amounts of unlabeled data. The initial approaches were based on the use of handcrafted learning objectives, which were actually grounded in the principle of invariance learning, whereas current SSL approaches are based on more principled learning objectives such as contrastive alignment and reconstruction. Modern objectives (contrastive/reconstruction) are more general-purpose and require less domain-specific engineering. Despite their success, most existing methods focus on aligning representations of observed data or reconstructing input signals, rather than modeling predictive structure. Again, some of the most successful current SSL methods are neither contrastive (using negative samples) nor reconstruction-based. This article presents the argument that predictive latent modeling represents the next generation of self-supervised learning evolution.\\
	
	\textbf{Structure of this paper:} The basic structure of this article comprises the following key elements:
	\begin{itemize}
		\item[i.] presents a unified taxonomy of SSL methods,
		\item[ii.] introduces Predictive Representation Learning as a new category,
		\item[iii.] positions JEPA as a canonical predictive SSL framework,
		\item[iv.] discusses theoretical implications from our empirical study and future research directions.
	\end{itemize}
	
	% =====================================================
	\section{Background and Fundamentals of Self-Supervised Learning}
	% =====================================================
	This section reviews the core principles of self-supervised learning, including representation collapse, architectural asymmetry, and learning signal design. We highlight the key dimensions along which SSL methods differ.
	
	% =====================================================
	\subsection{Alignment-Based Self-Supervised Learning}
	% =====================================================
    \textbf{Contrastive Learning:}
	Contrastive learning is a prominent paradigm in self-supervised learning that focuses on learning discriminative representations by comparing samples within a training batch. The central objective is to maximize the similarity between representations of different augmented views of the same input, known as positive pairs, while minimizing the similarity between representations of different inputs treated as negative pairs. This objective is commonly formulated using contrastive loss functions such as the InfoNCE loss\cite{rlcp}. Representative methods including SimCLR \cite{simclr} and Momentum Contrast (MoCo) \cite{moco} have demonstrated that contrastive objectives can produce high-quality representations that are competitive with supervised learning on downstream tasks \cite{simclr},\cite{moco}
	
	Despite their effectiveness, contrastive methods typically rely on large batch sizes or memory queues to provide sufficient negative samples, and their performance is sensitive to the choice of data augmentations and sampling strategies. These practical constraints increase computational complexity and motivate the development of alternative self-supervised learning approaches.\\
    
    \textbf{Non-Contrastive Alignment Methods:}
	The class of non-contrastive alignment algorithms is a type of self-supervised learning algorithm in which representations are learned by aligning the representations of different transformations of the same input but without using any negative examples explicitly. Rather than forcing differentiation between data points, non-contrastive algorithms ensure that no collapse occurs by making specific architecture choices, such as asymmetric networks, predictor heads, and stop gradients. The BYOL paper first proposed this concept by showing that a student network can predict its target network’s representation without negative pairs \cite{byol}. SimSiam further simplified this approach by removing the momentum encoder while retaining a predictor and stop-gradient mechanism \cite{simsiam}. However, despite the efficiency that is gained by not employing contrastive learning, which helps to greatly minimize the computational cost of training as well as the requirement of using large batch sizes and memory banks, the learning target still revolves around aligning representations between views.
	
	% =====================================================
	\subsection{Reconstruction-Based Self-Supervised Learning}
	% =====================================================
	Reconstruction self-supervised learning techniques learn representations through reconstruction of the input signal's missing/corrupted parts. Self-supervised learning is thus introduced by purposefully corrupting the input signal such that the representation learned can recover the signal. Classical self-supervised representation learning was pioneered by autoencoder methods, which have seen substantial development with the introduction of masking in recent years. For example, Masked AutoEncoders (MAE) work on the premise of masking a substantial portion of the input patches and reconstructing them. \cite{mae}. Similarly, methods such as BEiT leverage token reconstruction objectives inspired by masked language modeling \cite{beit}. Although reconstruction-based SSL effectively exploits partial observability and has demonstrated strong empirical results, these methods operate in input space, which imposes a high-dimensional reconstruction burden and may bias learning toward low-level details rather than semantic abstraction. These limitations motivate alternative approaches that shift the learning objective from input reconstruction to prediction in representation space.
	
	% =====================================================
	\subsection{Predictive Representation Learning}
	% =====================================================
	
	With recent developments in SSL, a new family of approaches is being discovered that is not only limited to learning invariance or reconstruction but also extends to prediction in representation space. Alignment-based models focus on learning invariance between multiple views, while reconstruction-based models restore the masked input through the learned representations. However, both approaches rely on exploiting the observations. The emerging prediction approach, however, focuses on learning using the unobserved/missing parts with the help of context. The increased focus on prediction indicates that there is an interest in exploring predictive structure in the data. This paper refers to this new family of models as PRL and treats them as a different class of SSL approaches rather than a newly proposed paradigm.\cite{jepa}.\\
	
	\textbf{Formal Definition:}
	Predictive Representation Learning encompasses self-supervised methods that define learning objectives in latent space through prediction of unobserved representations. Formally, given an unlabeled sample $x \sim \mathcal{D}$, the data is partitioned into an observed context $c(x)$ and an unobserved target $t(x)$. A context encoder maps $c(x)$ to a latent representation, while a target encoder produces a representation of $t(x)$. \rev{Training then minimizes a distance metric between the predicted and target embeddings, often using a separate target encoder.} Crucially, this formulation avoids both explicit negative sampling and reconstruction of the original input, distinguishing PRL from contrastive and generative SSL approaches \cite{jepa}.
	
	PRL methods learn representations by predicting the latent embedding of $t(x)$ from $c(x)$.\\
	
	\textbf{Core Properties}
	Predictive Representation Learning methods share several defining properties. 
	\begin{itemize}
		\item[i.] Supervision arises from latent-space prediction rather than instance discrimination or input reconstruction,
		\item[ii.] The learning objective introduces directionality, as representations of unobserved components are predicted from observed context,
		\item[iii.] Representational collapse is mitigated through architectural mechanisms such as asymmetric encoders, predictor networks, and stop-gradient or momentum updates, rather than contrastive negatives or explicit variance regularization,
		\item[iv.] By operating in representation space and emphasizing prediction under partial observability, PRL methods are well suited for scalable learning across vision, video, audio, and multimodal domains
	\end{itemize}
	
	In this study, we categorize JEPA-style architectures as representative implementations of PRL within this taxonomy.
	
	% =====================================================
	\section{Joint Embedding Predictive Architectures}
	% =====================================================
	\subsection{Architecture Overview}

	Joint Embedding Predictive Architectures (JEPA) belong to a class of predictive self-supervised learning methods that learn representations by predicting latent embeddings of unobserved data components from observed context. Unlike alignment-based or reconstruction-based approaches, JEPA does not compare multiple views symmetrically or reconstruct input signals. Instead, it formulates learning as a directional prediction problem in representation space.
	
	A typical JEPA architecture consists of three main components:
	\begin{itemize}
		\item[i.] \textbf{Context Encoder}: Maps the observed portion of an input sample to a latent representation.\\
		\item[ii.] \textbf{Target Encoder}: Produces a latent representation of an unobserved or masked portion of the input; its parameters are not directly updated by gradients.\\
		\item[iii.] \textbf{Predictor Network}: Transforms the context representation into a prediction of the target representation.
	\end{itemize}
	
	Architectural asymmetry between the context and target paths, often implemented using stop-gradient or momentum-based updates, plays a central role in preventing representational collapse.
	
	\subsection{Learning Objective}
	
	Let $x \sim \mathcal{D}$ denote an unlabeled data sample drawn from distribution $\mathcal{D}$.  
	The input is partitioned into:
	\begin{itemize}
		\item[i.] an observed context $c(x)$,
		\item[ii.] an unobserved target component $t(x)$.
	\end{itemize}
	
	The context encoder $f_{\theta}$ produces a latent representation:
	\begin{equation}
		z_c = f_{\theta}\big(c(x)\big).
	\end{equation}
	
	The target encoder $f_{\bar{\theta}}$, whose parameters are updated using stop-gradient or exponential moving average (EMA), produces:
	\begin{equation}
		z_t = f_{\bar{\theta}}\big(t(x)\big).
	\end{equation}
	
	A predictor network $g_{\phi}$ estimates the target representation:
	\begin{equation}
		\hat{z}_t = g_{\phi}(z_c).
	\end{equation}
	
	The JEPA training objective minimizes the discrepancy between predicted and target representations:
	\begin{equation}
		\mathcal{L}_{\text{JEPA}} =
		\mathbb{E}_{x \sim \mathcal{D}}
		\left[
		\left\|
		g_{\phi}\big(f_{\theta}(c(x))\big)
		-
		\mathrm{sg}\big(f_{\bar{\theta}}(t(x))\big)
		\right\|_2^2
		\right],
	\end{equation}
	where $\mathrm{sg}(\cdot)$ denotes the stop-gradient operation.
	
	This formulation defines learning entirely in latent space and avoids both explicit negative samples and input-level reconstruction.

	% =====================================================
	\section{Comparative Analysis of SSL Paradigms}
	% =====================================================
	This section compares alignment-based, reconstruction-based, and predictive SSL methods in terms of objectives, collapse avoidance, and scalability.
	
	\subsection {Architectural Comparison}
\begin{figure}[htbp] % Optional placement specifier
		\centering       % Center the image
		\includegraphics[width=1\textwidth]{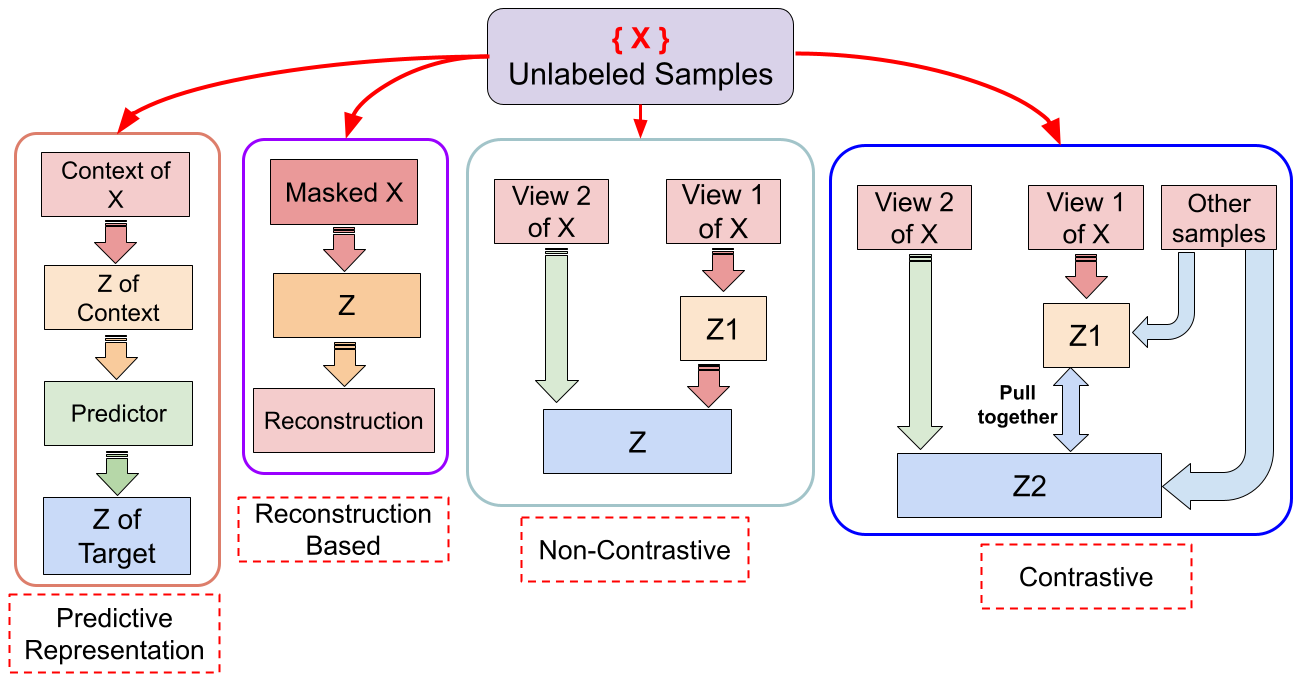} % Adjust the width as needed
		\caption{Architectural comparison of Contrastive, Non Contrastive, Reconstruction based and Predictive Representation} % Optional caption for the figure
		\label{fig:architecture_comparison}  % Optional label for referencing
	\end{figure}

	As shown in Fig.~\ref{fig:architecture_comparison}, contrastive and non-contrastive self-supervised learning methods rely on symmetric view-based architectures, with contrastive approaches additionally requiring negative samples or memory queues. Reconstruction-based methods employ encoder–decoder designs that operate in input space to recover masked signals. In contrast, JEPA adopts an asymmetric dual-path architecture that predicts latent representations of unobserved components from context, avoiding both negative samples and input reconstruction.
	
	\subsection{Representative Loss Functions and Formulations}
	
	\subsubsection{Contrastive Loss (InfoNCE)}
	
	Contrastive loss functions are widely used in methods such as SimCLR and MoCo to learn discriminative representations. A commonly adopted formulation is the InfoNCE loss:
	
	\begin{equation}
		\mathcal{L}_{\text{InfoNCE}} =
		- \log
		\frac{
			\exp\left(\mathrm{sim}(z_i, z_i^+)/\tau\right)
		}{
			\exp\left(\mathrm{sim}(z_i, z_i^+)/\tau\right)
			+ \sum_{j \neq i} \exp\left(\mathrm{sim}(z_i, z_j)/\tau\right)
		},
	\end{equation}
	
	where $z_i$ and $z_i^+$ denote representations of a positive pair, $z_j$ denotes representations of negative samples, $\mathrm{sim}(\cdot,\cdot)$ is a similarity measure such as cosine similarity, and $\tau$ is a temperature parameter.
	
	This loss prevents representational collapse through the use of negative samples and encourages instance-level discrimination.\\
	
	\textbf{Non-Contrastive Alignment Loss (BYOL / SimSiam):}
	
	Non-contrastive alignment methods eliminate negative samples and rely on architectural asymmetry and gradient blocking. A typical loss function is defined as:
	
	\begin{equation}
		\mathcal{L}_{\text{NC}} =
		\left\|
		g_{\phi}\big(f_{\theta}(x_1)\big)
		-
		\mathrm{sg}\big(f_{\bar{\theta}}(x_2)\big)
		\right\|_2^2,
	\end{equation}
	
	where $f_{\theta}$ is the online encoder, $f_{\bar{\theta}}$ is the target encoder updated using stop-gradient or momentum mechanisms, $g_{\phi}$ is a predictor network, and $\mathrm{sg}(\cdot)$ denotes the stop-gradient operation.
	
	These methods avoid collapse through predictor asymmetry and gradient blocking, while the objective enforces symmetric alignment between representations.
	
	\subsubsection{Reconstruction Loss (Autoencoders / MAE)}
	
	Reconstruction-based self-supervised learning methods derive supervision by recovering masked or corrupted inputs. A generic reconstruction loss can be expressed as:
	
	\begin{equation}
		\mathcal{L}_{\text{rec}} =
		\left\|
		x - D\big(E(\tilde{x})\big)
		\right\|_2^2,
	\end{equation}
	
	where $\tilde{x}$ represents a masked or corrupted version of the input $x$, $E(\cdot)$ is an encoder, and $D(\cdot)$ is a decoder.
	
	Collapse is avoided through reconstruction fidelity; however, these losses operate in input space and are sensitive to low-level signal details.
	
	\subsubsection{Predictive Representation Learning Loss (JEPA)}
	
	Predictive Representation Learning methods, including JEPA-style architectures, define learning objectives in latent space through prediction of unobserved components. A representative loss is given by:
	
	\begin{equation}
		\mathcal{L}_{\text{PRL}} =
		\mathbb{E}_{x \sim \mathcal{D}}
		\left[
		\left\|
		g_{\phi}\big(f_{\theta}(c(x))\big)
		-
		\mathrm{sg}\big(f_{\bar{\theta}}(t(x))\big)
		\right\|_2^2
		\right],
	\end{equation}
	
	where $c(x)$ denotes the observed context, $t(x)$ the unobserved target component, and the remaining terms follow standard notation.
	
	This loss predicts latent representations of unseen components, does not rely on negative samples or input reconstruction, and avoids collapse through predictive structure.

	Comparison is shown in Table.~\ref{tab:ssl_losses}.
    \rev{
While both Table 1 and Table 4 compare self-supervised learning paradigms, they serve complementary purposes. Table 1 provides a high-level conceptual and architectural comparison across methods, whereas Table 4 focuses specifically on loss functions and optimization characteristics, including objective formulation and collapse avoidance mechanisms.
}

	\begin{table}[t]
		\centering
		\caption{\rev{Conceptual and architectural comparison}}
		\label{tab:ssl_comparison}
		\resizebox{\textwidth}{!}{
			\begin{tabular}{lcccc}
				\toprule
				\textbf{Dimension} &
				\textbf{Contrastive SSL} &
				\textbf{Non-Contrastive SSL} &
				\textbf{Reconstruction SSL } &
				\textbf{ Predictive Representation Learning} \\
				\midrule
				Representative methods &
				SimCLR, MoCo &
				BYOL, SimSiam &
				MAE, BEiT &
				JEPA-style methods \\
				
				Learning signal &
				Discrimination &
				View alignment &
				Input reconstruction &
				Latent prediction \\
				
				Negative samples &
				Yes &
				No &
				No &
				No \\
				
				Operates in input space &
				No &
				No &
				Yes &
				No \\
				
				Operates in latent space &
				Yes &
				Yes &
				Partially &
				Yes \\
				
				Objective symmetry &
				Symmetric &
				Symmetric &
				Directional &
				Directional \\
				
				Prediction target &
				Instance embeddings &
				View embeddings &
				Pixels / tokens &
				Unseen representations \\
				
				Collapse prevention &
				Negatives &
				Architecture, stop-gradient &
				Reconstruction loss &
				Architectural prediction \\
				
				Computational cost &
				High &
				Moderate &
				High &
				Moderate \\
				
				Semantic abstraction &
				Medium &
				Medium &
				Low--medium &
				High \\
				
				World-model capability &
				Weak &
				Weak &
				Limited &
				Strong \\
				\bottomrule
			\end{tabular}
		}
	\end{table}

    %====== Different implementation of JEPA ========
    
    \subsection{\rev{Empirical Comparison: Alignment vs Reconstruction vs Predictive Learning}}

\rev{
To complement the conceptual analysis, we present an empirical comparison across three major self-supervised learning paradigms: alignment-based (BYOL), reconstruction-based (MAE), and predictive representation learning (I-JEPA).
}\\

\rev{
\textbf{Evaluation Metrics:}
\begin{itemize}
    \item[i.] Augmentation Similarity (mean $\pm$ std)
    \item[ii.] Occlusion Robustness (mean $\pm$ std)
\end{itemize}
}

\begin{table}[h]
\centering
\caption{\rev{Empirical Comparison of SSL Paradigms from our implementation and training results of BYOL, I-JEPA  for their Allighment, PRL and MAE}}
{\setlength{\tabcolsep}{10pt}%
\begin{tabular}{lcc}
\hline
Method & Augmentation Similarity & Occlusion Robustness  \\
\hline
BYOL (Alignment) & $0.988 \pm 0.007$ & $0.750 \pm 0.046$ \\
I-JEPA (PRL) & $0.950 \pm 0.027$ & $\mathbf{0.788 \pm 0.045}$ \\
MAE (Reconstruction) & $\mathbf{1.000 \pm 0.000}$ & $0.550 \pm 0.086$ \\
\hline
\end{tabular}}
\end{table}

\rev{
\textbf{Observations:} The following observations are derived from the empirical evaluation of BYOL, I-JEPA, and MAE based on our implementation.
\begin{itemize}
    \item [i.] MAE achieves perfect similarity due to pixel-level reconstruction, but exhibits significantly lower robustness.
    \item[ii.]  BYOL achieves high similarity through strong alignment, but its robustness is lower than predictive methods.
    \item[iii.]  I-JEPA achieves the best robustness, demonstrating superior ability to handle occlusion and partial observability.
\end{itemize}
}

\rev{
\textbf{Key Insight:}
These results highlight a fundamental trade-off: alignment and reconstruction objectives optimize similarity, whereas predictive objectives improve robustness. Predictive Representation Learning (PRL) provides a better balance by capturing structural dependencies, leading to improved generalization.
}
\begin{figure}
\centering
\includegraphics[width=0.98\linewidth]{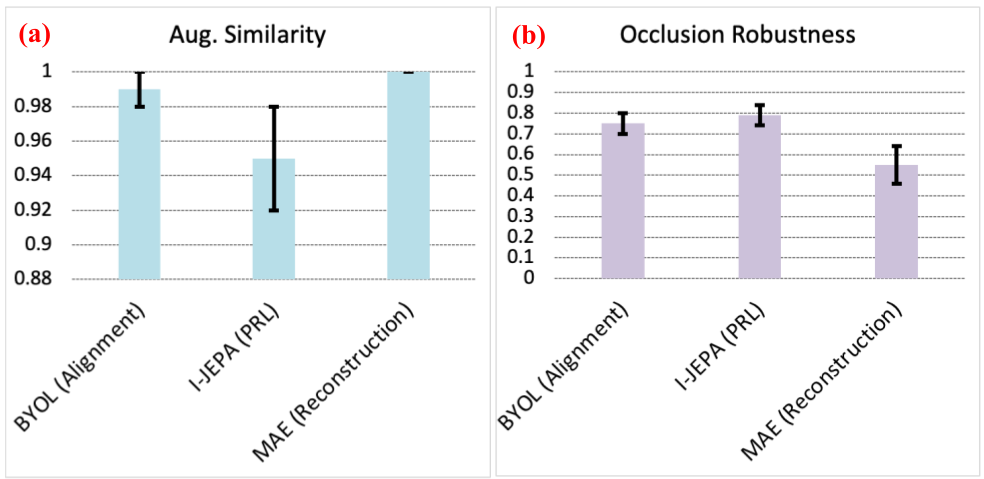}
\caption{\rev{Comparision of the results from our custom implimentation and training: Comparison of augmentation similarity and occlusion robustness across BYOL, I-JEPA, and MAE. Predictive learning (I-JEPA) achieves superior robustness despite lower similarity.}}
\label{fig:ssl_comparison}
\end{figure}
    \subsection{\rev{Benchmark Results of JEPA Variants}}

\rev{
We summarize representative benchmark results from published JEPA-based models across modalities. These results are drawn from recent CVPR and ICLR works and demonstrate the effectiveness of predictive representation learning.
}

\begin{table}[h]
\centering
\caption{\rev{Published Benchmark Results for JEPA Variants}}
\resizebox{\textwidth}{!}{%
\begin{tabular}{lcccc}
\hline
Model & Modality & Dataset & Evaluation Protocol & Performance \\
\hline
I-JEPA~\cite{ijepa} 
& Vision 
& ImageNet-1K 
& Linear Probe (ViT-H/14) 
& 72.8\% Top-1 \\

I-JEPA~\cite{ijepa} 
& Vision 
& ImageNet-1K 
& k-NN 
& $\sim$73\% \\

V-JEPA~\cite{bardes2024vjepa} 
& Video 
& Kinetics-400 
& Linear Probe 
& Competitive with VideoMAE \\

VL-JEPA~\cite{vl-jepa} 
& Vision-Language 
& Image-Text datasets 
& Retrieval / Alignment 
& Improved cross-modal retrieval \\

Graph-JEPA~\cite{graph-jepa} 
& Graph 
& OGB datasets 
& Node Classification 
& SOTA / Competitive \\
\hline
\end{tabular}
}
\end{table}

\rev{
\textbf{Observations:} From both the benchmark results (published till date) and from our custom implementation results, we may  draw the following observations:
\begin{itemize}
    \item [i.] I-JEPA achieves strong performance on ImageNet linear evaluation without relying on reconstruction or contrastive losses.
    \item [ii.] V-JEPA demonstrates competitive performance with reconstruction-based video SSL methods such as VideoMAE.
    \item [iii.] Multimodal and graph JEPA variants show that predictive objectives generalize across domains.
    \item [iv.] These results highlight that predictive representation learning can achieve competitive performance while operating entirely in latent space.
\end{itemize}
}
    \subsection{Implementations of Joint Embedding Predictive Architectures}

Joint Embedding Predictive Architectures have been instantiated across multiple data modalities and learning settings, demonstrating the generality of predictive representation learning.

\begin{itemize}

  \item [i.]\textbf{I-JEPA (Image JEPA)}:
  A non-generative self-supervised learning approach that predicts latent embeddings of masked image regions from observed context, enabling the learning of semantic visual representations without pixel-level reconstruction~\cite{ijepa}.

  \item[ii.] \textbf{VL-JEPA (Vision--Language JEPA)}:
  Extends the JEPA framework to jointly model visual and textual modalities by predicting continuous language embeddings from visual context, supporting unified vision--language representation learning~\cite{vl-jepa}.

  \item[iii.] \textbf{V-JEPA 2 (Video JEPA)}:
  A video-based extension of JEPA that predicts representations of future or masked spatio-temporal regions, facilitating improved modeling of temporal dynamics and long-range dependencies~\cite{vjepa}.

  \item[iv.]\textbf{graph-JEPA}:
  Applies JEPA principles to graph representation learning by predicting embeddings of nodes or subgraphs from observed graph context, enabling predictive learning on structured, non-Euclidean data~\cite{graph-jepa}.

  \item[v.] \textbf{d-jepa}:Integrates JEPA-style latent prediction with denoising and generative modeling perspectives, interpreting JEPA as a generalized prediction framework in continuous representation spaces~\cite{d-jepa}.

  \item[vi.] \textbf{deq-jepa}:
  A sequence-oriented JEPA variant that performs discriminative sequential prediction of representations, enabling richer modeling of ordered visual or temporal data~\cite{dseq-jepa}.

  \item[vii.] \textbf{seq-jepa}:
  Combines JEPA with autoregressive and equivariant representation learning to support world modeling and structured prediction in sequential environments~\cite{seq-jepa}.

\end{itemize}

	%==============New Taxonomy ==================

	\section{New Taxonomy of Self-Supervised Learning}
	
\begin{figure}[htbp] % Optional placement specifier
		\centering       % Center the image
		\includegraphics[width=1\textwidth]{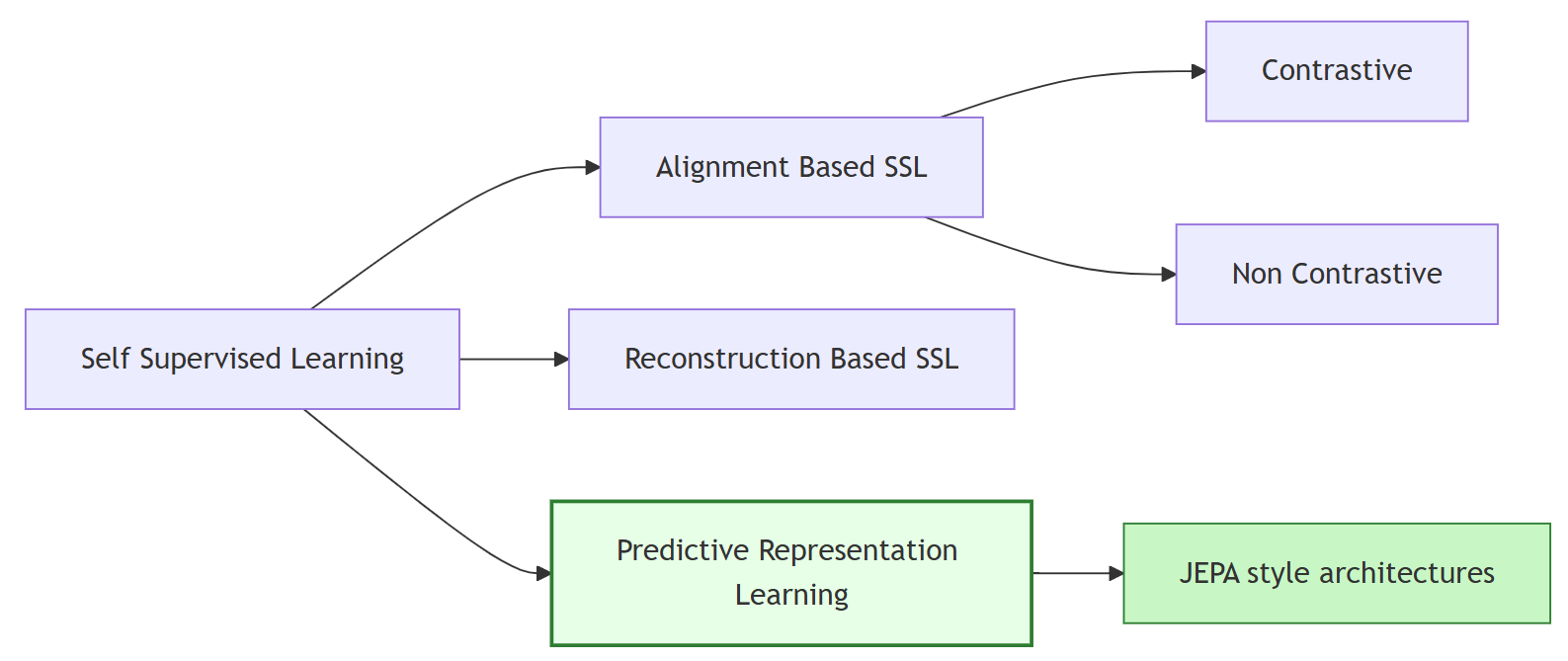} % Adjust the width as needed
		\caption{New Taxonomy for SSL categorization} % Optional caption for the figure
		\label{fig:taxonomy}  % Optional label for referencing
	\end{figure}

	This study adopts a taxonomy that organizes self-supervised learning methods according to the nature of their learning objectives rather than specific model architectures or data modalities, as illustrated in Fig.~\ref{fig:taxonomy}. Existing approaches are categorized into alignment-based methods, reconstruction-based methods, and predictive representation learning methods. Alignment-based approaches derive supervision by matching representations of observed views, reconstruction-based approaches focus on recovering masked input signals, while predictive representation learning emphasizes latent-space prediction of unobserved components from context. This taxonomy clarifies conceptual differences across self-supervised learning paradigms and provides a structured framework for positioning predictive architectures such as JEPA within the broader SSL landscape. The conceptual differences are listed in Table.~\ref{tab:ssl_comparison}

	\begin{table}[t]
		\centering
		\caption{\rev{Loss functions and optimization characteristics}}
		\label{tab:ssl_losses}
		\resizebox{\textwidth}{!}{
			\begin{tabular}{lcccc}
				\toprule
				\textbf{Aspect} &
				\textbf{Contrastive} &
				\textbf{Non-Contrastive} &
				\textbf{Reconstruction} &
				\textbf{Predictive (PRL)} \\
				\midrule
				Representative loss &
				InfoNCE &
				Alignment loss &
				Reconstruction loss &
				Latent prediction loss \\
				
				Learning signal &
				Discrimination &
				View matching &
				Input recovery &
				Context-to-target prediction \\
				
				Negative samples &
				Required &
				Not required &
				Not required &
				Not required \\
				
				Operates in input space &
				No &
				No &
				Yes &
				No \\
				
				Operates in latent space &
				Yes &
				Yes &
				Partially &
				Yes \\
				
				Objective symmetry &
				Symmetric &
				Symmetric &
				Directional &
				Directional \\
				
				Collapse avoidance &
				Negatives &
				Architecture, stop-grad &
				Reconstruction fidelity &
				Predictive inconsistency \\
				
				Sensitivity to low-level details &
				Low &
				Low &
				High &
				Low \\
				
				Encouraged representation &
				Instance separation &
				Invariance &
				Local detail &
				Structural dependency \\
				\bottomrule
			\end{tabular}
		}
	\end{table}

	%=========Learning Signal Comparison  ==========

	% =====================================================
	\section{Why Predictive Representation Learning Is a Distinct SSL Category}
	% =====================================================
	Predictive Representation Learning (PRL) is best understood as a distinct category within self-supervised learning due to fundamental differences in its learning objective, supervision mechanism, and representational focus. Rather than extending existing alignment or reconstruction strategies, PRL reframes self-supervision around prediction of latent representations corresponding to unobserved data, leading to qualitative differences in how representations are learned.
	\subsection{Learning Objective Perspective}
	Most alignment-based self-supervised methods whether contrastive or non-contrastive optimize objectives that enforce consistency between representations derived from different views of the same observed input. Reconstruction-based methods instead minimize discrepancies between original inputs and their reconstructed versions. In both cases, optimization is driven by signals derived from data that is directly available to the model.
	
	PRL departs from this formulation by defining learning as a prediction task in representation space, where the objective is to estimate the embedding of a data component that is not directly observed. This predictive formulation introduces directionality and shifts the emphasis from similarity or fidelity to anticipation of missing information.
	\subsection{Nature of Self-Supervision}
	In contrastive learning, supervision arises from distinguishing between positive and negative sample pairs, while non-contrastive alignment methods rely on architectural constraints to match representations across views. Reconstruction-based approaches generate supervision by recovering masked or corrupted input signals.
	
	In PRL, the supervisory signal is derived from the consistency between predicted and target embeddings of unobserved data components. Because the target representation corresponds to information outside the model’s immediate input, learning is driven by predictive relationships rather than surface-level similarity or reconstruction accuracy.
	\subsection{Role of Partial Observability}
	While reconstruction-based methods introduce partial observability through masking, their objective ultimately requires exact recovery of missing input elements. Alignment-based methods generally assume that each view provides a complete observation of the underlying sample.
	
	PRL treats partial observability as an inherent condition of learning rather than a temporary obstacle. The model is not required to recover missing inputs, but only to predict their representations, allowing multiple possible inputs to map to similar latent targets. This encourages abstraction and reduces sensitivity to low-level variations.
	
	\subsection{Collapse Avoidance Mechanisms}
	
	A central challenge in self-supervised learning is avoiding representational collapse, where all inputs are mapped to identical or low-variance representations. Different self-supervised learning paradigms address this issue through distinct objective formulations and architectural constraints. This section summarizes the collapse avoidance mechanisms used in alignment-based, reconstruction-based, and predictive representation learning approaches.
	
	\subsubsection{Contrastive Self-Supervised Learning}
	
	Contrastive methods prevent collapse by explicitly enforcing separation between representations of different samples. A commonly used objective is the InfoNCE loss~\cite{rlcp}:
	
	\begin{equation}
		\mathcal{L}_{\text{InfoNCE}} =
		- \log
		\frac{
			\exp\left(\mathrm{sim}(z_i, z_i^+)/\tau\right)
		}{
			\exp\left(\mathrm{sim}(z_i, z_i^+)/\tau\right)
			+ \sum_{j \neq i} \exp\left(\mathrm{sim}(z_i, z_j)/\tau\right)
		},
	\end{equation}
	
	where $z_i$ and $z_i^+$ denote representations of positive pairs, $z_j$ denotes representations of negative samples, $\mathrm{sim}(\cdot,\cdot)$ is cosine similarity, and $\tau$ is a temperature parameter. Collapsed representations yield identical similarities for positive and negative pairs, resulting in a high loss and thus being penalized during optimization~\cite{simclr,moco}.
	
	\subsubsection{Non-Contrastive Alignment Methods}
	
	Non-contrastive alignment methods avoid collapse without relying on negative samples by introducing architectural asymmetry and gradient blocking. A typical objective used in methods such as BYOL and SimSiam is:
	
	\begin{equation}
		\mathcal{L}_{\text{NC}} =
		\left\|
		g_\phi(f_\theta(x_1)) -
		\mathrm{sg}\left(f_{\bar{\theta}}(x_2)\right)
		\right\|_2^2,
	\end{equation}
	
	where $f_\theta$ denotes the online encoder, $f_{\bar{\theta}}$ the target encoder updated via stop-gradient or exponential moving average, $g_\phi$ a predictor network, and $\mathrm{sg}(\cdot)$ denotes the stop-gradient operation. The asymmetry introduced by the predictor and gradient blocking prevents trivial constant solutions from becoming stable optima~\cite{byol,simsiam}.
	
	\subsubsection{Reconstruction-Based Self-Supervised Learning}
	
	Reconstruction-based methods implicitly prevent collapse by enforcing fidelity to the input signal. A generic reconstruction loss can be expressed as:
	
	\begin{equation}
		\mathcal{L}_{\text{rec}} =
		\left\|
		x - D(E(\tilde{x}))
		\right\|^2,
	\end{equation}
	
	where $\tilde{x}$ is a masked or corrupted version of the input $x$, $E(\cdot)$ is an encoder, and $D(\cdot)$ is a decoder. Collapsed representations are insufficient to reconstruct diverse inputs, resulting in large reconstruction error~\cite{mae,beit}.
	
	\subsubsection{Predictive Representation Learning}
	
	Predictive Representation Learning (PRL) avoids collapse by defining learning as prediction of latent representations corresponding to unobserved data components. A general PRL objective can be written as:
	
	\begin{equation}
		\mathcal{L}_{\text{PRL}} =
		\mathbb{E}_{x \sim \mathcal{D}}
		\left[
		\left\|
		g_\phi\big(f_\theta(c(x))\big)
		-
		\mathrm{sg}\big(f_{\bar{\theta}}(t(x))\big)
		\right\|_2^2
		\right],
	\end{equation}
	
	where $c(x)$ denotes an observed context, $t(x)$ an unobserved target component, $f_\theta$ and $f_{\bar{\theta}}$ are context and target encoders respectively, and $g_\phi$ is a predictor. In this setting, collapsed representations are unable to predict diverse unseen targets across samples, leading to high prediction error. Collapse is therefore avoided through the predictive structure of the objective rather than explicit negatives or reconstruction losses~\cite{jepa}.
	\subsection {Representational Emphasis}
	Alignment-based approaches primarily learn invariances to data augmentations, while reconstruction-based methods often retain fine-grained details necessary for input recovery. In contrast, PRL emphasizes representations that encode structural and relational information, as effective prediction of unobserved components requires capturing dependencies across different parts of the data.
	
	\begin{table}[t]
		\centering
		\caption{Conceptual differences between self-supervised learning categories}
		\label{tab:prl_vs_ssl}
		\resizebox{\textwidth}{!}{
			\begin{tabular}{lccc}
				\toprule
				\textbf{Dimension} &
				\textbf{Alignment-Based SSL} &
				\textbf{Reconstruction-Based SSL} &
				\textbf{Predictive Representation Learning} \\
				\midrule
				Primary learning objective &
				Representation alignment &
				Input recovery &
				Latent prediction \\
				
				Source of supervision &
				Observed views &
				Observed inputs &
				Unobserved components \\
				
				Objective formulation &
				Symmetric &
				Directional &
				Directional \\
				
				Learning space &
				Latent space &
				Input space &
				Latent space \\
				
				Role of masking or augmentation &
				View generation &
				Exact reconstruction &
				Partial observability \\
				
				Treatment of missing information &
				Implicitly ignored &
				Explicitly recovered &
				Predicted in representation space \\
				
				Collapse avoidance mechanism &
				Negatives or architectural constraints &
				Reconstruction loss &
				Predictive structure \\
				
				Representational focus &
				Invariance to transformations &
				Local signal fidelity &
				Structural dependencies \\
				
				Sensitivity to low-level details &
				Low &
				High &
				Low \\
				
				Suitability for world modeling &
				Limited &
				Limited &
				Strong \\
				\bottomrule
			\end{tabular}
		}
	\end{table}

	% =====================================================
	\section{Open Challenges and Future Directions of Joint Embedding Predictive Architectures (JEPA)}
    Based on this study we find the following challenges for JEPA architecture: 
	
	\begin{itemize}
		\item[i.] \textbf{Theoretical Understanding of Predictive Objectives}:  
		A key open challenge is the lack of formal theoretical guarantees explaining why latent predictive objectives yield stable and informative representations, particularly in the absence of contrastive negatives or reconstruction losses.
		
		\item[ii.] \textbf{Long-Horizon Prediction and Temporal Abstraction}:  
		Extending JEPA to long-term temporal prediction remains challenging due to issues such as error accumulation, representation drift, and the need for hierarchical temporal abstraction.
		
		\item[iii.] \textbf{Multimodal and Cross-Modal Prediction}:  
		While JEPA naturally supports multimodal learning, designing effective cross-modal predictive objectives and balancing modality-specific and shared representations remain open problems.
		
		\item[iv.] \textbf{Scalability and Architectural Design Choices}:  
		Understanding how model size, predictor capacity, masking strategies, and encoder asymmetry affect performance is an open challenge, particularly for large-scale deployment.
		
		\item[v.] \textbf{Evaluation and Benchmarking}:  
		Current evaluation protocols emphasize downstream task performance, which may not fully reflect predictive capability. Developing benchmarks that directly assess latent prediction quality is an important future direction.
		
		\item[vi.] \textbf{Robustness Under Partial Observability}:  
		Although JEPA is designed for partial observability, systematic evaluation of robustness to missing, noisy, or ambiguous context remains limited and warrants further study.
		
		\item[vii.] \textbf{Integration with Embodied and Interactive Systems}:  
		Applying JEPA to embodied agents and interactive environments introduces challenges related to online learning, exploration, and coupling predictive representations with control.
		
		\item[viii.] \textbf{Connections to Cognitive and Theoretical Frameworks}:  
		Clarifying the relationship between JEPA and predictive coding, energy-based models, and free-energy minimization remains an open area for interdisciplinary research.
	\end{itemize}

	% =====================================================
	\section{Conclusion}
	% =====================================================
	In this work, we propose a taxonomic analysis of emerging self-supervised learning frameworks in terms of their objective functions and the signals used to train them. Namely, alignment-based and reconstruction-based approaches are considered from a single perspective, enabling a comparative evaluation of their representational capabilities, bias-inducing priors, and weaknesses in modeling the observed data distribution. In the proposed taxonomy, Predictive Representation Learning (PRL) approach is described as an independent class of methods, where optimization targets for latent representations are determined via predictive objective functions when only a limited part of the data is observable. In this setting, Joint Embedding Predictive Architectures (JEPA) are introduced as a set of canonical models, which implement predictive consistency of latent representations conditioned on contextual features and targets' representations. We also discusses the differences in the structure of predictive objectives from that of alignment or reconstruction losses, especially with respect to information efficiency, invariant transformations, and implicit regularizers.

	% =====================================================
	% References
	% =====================================================
	\sloppy

\end{document}